# MAPCHANGE: ENHANCING SEMANTIC CHANGE DETECTION WITH TEMPORAL-INVARIANT HISTORICAL MAPS BASED ON DEEP TRIPLET NETWORK

*Yinhe Liu[1], Sunan Shi[1], Zhuo Zheng[2], Jue Wang[1], Shiqi Tian[1], Yanfei Zhong[1, *], Senior Member, IEEE*

1) State Key Laboratory of Information Engineering in Surveying, Mapping, and Remote Sensing, Wuhan University, P. R. China
2) Department of Computer Science, Stanford University

## ABSTRACT

Semantic Change Detection (SCD) is recognized as both a crucial and challenging task in the field of image analysis. Traditional methods for SCD have predominantly relied on the comparison of image pairs. However, this approach is significantly hindered by substantial imaging differences, which arise due to variations in shooting times, atmospheric conditions, and angles. Such discrepancies lead to two primary issues: the under-detection of minor yet significant changes, and the generation of false alarms due to temporal variances. These factors often result in unchanged objects appearing markedly different in multi-temporal images. In response to these challenges, the MapChange framework has been developed. This framework introduces a novel paradigm that synergizes temporal-invariant historical map data with contemporary high-resolution images. By employing this combination, the temporal variance inherent in conventional image pair comparisons is effectively mitigated. The efficacy of the MapChange framework has been empirically validated through comprehensive testing on two public datasets. These tests have demonstrated the framework's marked superiority over existing state-of-the-art SCD methods.

## 1. INTRODUCTION

In recent years, a remarkable advancement has been the surge in availability of remote sensing images with high spatial-temporal resolution. This development has facilitated more sophisticated analytical techniques, including in change detection (CD) tasks, which can be categorized into binary CD (BCD) and semantic CD (SCD) [1]. BCD focuses on identifying the locations where changes have occurred, whereas SCD goes a step further by discerning both the locations and the nature of these changes. The relevance of SCD spans across various critical sectors, including urban planning, environmental monitoring, disaster management, and agricultural assessment [2].

In advancing SCD methodologies, the Post-Classification Comparison (PCC) method has been a primary approach [2]. This method views SCD as a dual temporal image classification problem, circumventing the classification space's sparsity in directly identifying semantic changes. PCC methods build upon single-temporal classification approaches by independently classifying each image and comparing these classification maps. While the simplicity of this method facilitates the integration of advanced classification techniques, it has limitations, particularly in neglecting the temporal relationship between image pairs, leading to frequent false alarms, especially near object edges. To overcome these challenges, the multi-task framework utilizing image pairs as inputs has been developed [3]. This framework not only includes a classification head for each image but also introduces a binary change detection head to refine semantic change outputs. The advent of Siamese Networks for the fusion of classification with binary change detection have significantly improved the effectiveness of multi-task SCD methods, showcasing continual progress in this area.

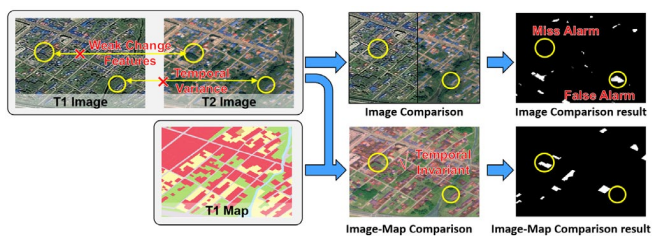

**Fig 1.** Illustration of image-comparison- and image-map comparison-based change detection methods.

As shown in **Fig. 1**, current SCD methods primarily depend on comparison of image pairs. However, this approach faces challenges due to significant imaging difference between two images affected by different shooting time, atmospheric radiation, shooting angle and other factors. These disparities often result in two issues: 1) overlooking minor but significant changes, leading to a lower recall rate, and 2) triggering false alarms due to *temporal-variance* problem, which means unchanged objects may have visually prominent changes in multi-temporal images. To solve the abovementioned problems, this paper utilizes historical maps or vector data which are often available in practical applications, to provide a rich source of semantic prior information that can greatly enhance SCD effectiveness. The semantic information in these historical maps is typically *temporal-invariant*, i.e., the unchanged objects are invariant over multi-temporal maps, offering a solid counterbalance to the temporal variances in image pairs. By incorporating these historical semantic layers, this paper aims to develop a more nuanced and accurate SCD approach, mitigating the limitations of temporal discrepancies.

The key contributions of this work are summarized as follows:

1) Introduction of a novel SCD paradigm, MapChange, which integrates temporal-invariant historical map data with current high-resolution images to reconcile temporal discrepancies in traditional image pair comparisons.

2) Development of a specialized triplet network architecture featuring distinct image and map encoders, enhancing the integration of multi-modal data sources.

3) Empirical demonstration of the MapChange framework's superiority through experiments on two public datasets, evidencing its significant advancement over existing state-of-the-art SCD methods.

## 2. METHOD

The MapChange framework presents an approach to semantic change detection, incorporating historical maps alongside temporal high-resolution images to assist in the identification of changes. As shown in **Fig. 1**, MapChange framework comprises a triple network architecture to handle the distinct data sources, integrates the information through a modal fusion process, and applies specific decoders for the tasks of semantic classification and change detection, aiming to improve the precision of SCD by leveraging the context provided by historical maps.

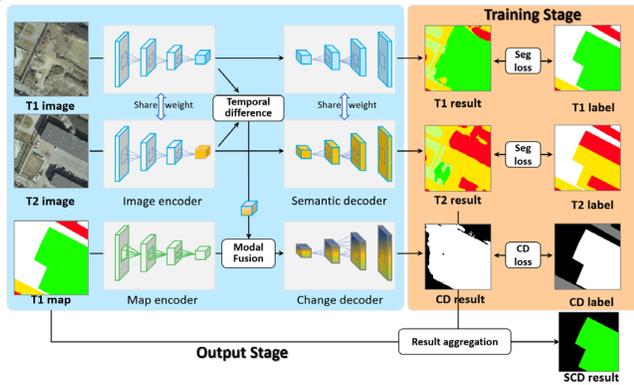

**Fig. 1.** Flowchart of the proposed historical map prompted semantic change detection framework MapChange.

### 2.1 Triple Network

The MapChange framework introduces a triplet network architecture which is distinguished from traditional Siamese networks by accommodating an additional input stream for historical maps. In this configuration, image pairs $I_{T1}$ and $I_{T2}$ are processed through a shared-weight image encoder $f(\cdot)$, ensuring temporal feature consistency. Concurrently, the historical map $M$ is represented in a one-hot encoded format and processed by a dedicated map encoder $g(\cdot)$ extract relevant historical features. The encoders can be mathematically represented as $E_{T1} = f(I_{T1})$, $E_{T2} = f(I_{T2})$, and $E_M = g(M)$, where $E_{T1}$, $E_{T2}$, and $E_M$ denote the feature maps of the first temporal (T1) image, the second temporal (T2) image, and the historical map corresponding to the T1 image, respectively. This architecture is designed to exploit the semantic priors offered by historical maps, enhancing the network's ability to discern semantic changes effectively.

### 2.2 Multi-modal feature fusion

Within the MapChange framework, multi-modal feature fusion is a pivotal step for enhance image change feature via historical map. Initially, the image features extracted by the encoders undergo a temporal difference operation to isolate change features, denoted as $C_{diff} = \nabla(E_{T1}, E_{T2})$ where $\nabla(\cdot)$ represents the temporal difference function that can involve addition, subtraction, or concatenation operations. In our approach, we employ a Temporal-Symmetric Transformer (TST) $\tau(\cdot)$ following the methodology of ChangeMask [3], designed to ensure temporal symmetry and maintain discriminative properties for the change representation $C_{TST} = \tau(C_{diff})$.

Subsequently, the change features are enhanced by the historical map features through fusion. The fusion process can be expressed as $C_{fused} = \phi(C_{TST}, E_M)$, where $\phi(\cdot)$ symbolizes the fusion function, which also experimented with TST and concatenation operations. Empirical results indicated that a straightforward concatenation $C_{fused} = [C_{TST}; E_M]$ provided the most effective performance for enhancing change features with historical map context, as detailed in Experiment Section. This multi-modal fusion strategy is instrumental in the MapChange framework, enriching the change detection process with valuable semantic priors from historical data.

### 2.3 Multi-task Decoders

The MapChange framework employs a multi-task decoder design, consisting of two semantic decoders with shared weights, denoted as $D_{sem}$, for semantic segmentation tasks, and an independent change decoder $D_{chg}$ for Binary Change Detection (BCD). At the inference stage, despite the semantic decoders' ability to produce classification maps for T1, the pre-existing T1 map is utilized to enhance the semantic change detection outcome. An end-to-end training scheme is adopted, with a multi-task loss function $L$, .where $L_{cls}^{t_1}$ and $L_{cls}^{t_2}$ are the cross-entropy losses supervising the semantic segmentation, and $L_{bcd}$ is a hybrid of dice loss $L_{dice}$ and

binary cross-entropy loss $L_{bce}$ for BCD. The combined loss function is formulated as:

$$L = L_{cls}^{t_1} + L_{cls}^{t_2} + L_{bcd} \quad (1)$$

## 3. EXPERIMENT

### 3.1 Experiment Setup
*3.1.1 Datasets*
In this study, two key datasets were employed: Hi-UCD [4] and HRSCD [1]. Both datasets offer unique challenges and are integral to evaluating the efficacy of semantic change detection algorithms.

The HRSCD dataset contains 291 pairs of aerial images (10000×10000 pixels, 50 cm/pixel resolution), captured in two phases (2005/2006 and 2012), and is annotated for both change and terrain. To manage the dataset's volume, only patches within change areas were selected, with a division of 70% for training and 30% for testing.

The Hi-UCD dataset, focusing on urban changes in Estonia, includes 745 pairs of images (0.1 m resolution, 1024×1024 pixels), with nine land-cover classes and a potential of 82 land-cover change types. The dataset split comprises 300 training pairs, 59 validation pairs, and 386 testing pairs, ensuring spatial independence across these sets.

*3.1.2 Implementation details and evaluation metrics*
In the experimental implementation, the model was trained using a batch size of 16. The optimization was conducted using Stochastic Gradient Descent (SGD), configured with a momentum of 0.9 and a weight decay of 0.0001. The learning rate followed a polynomial decay schedule, starting from a base learning rate of 0.03. The training iterations were set to 15,000 for the HiUCD dataset and 60,000 for the HRSCD dataset. All experiments were conducted on a computational setup equipped with two NVIDIA TITAN RTX graphics cards.

In evaluating the performance of semantic change detection (SCD) models in this study, a comprehensive set of metrics was employed. These include Overall Accuracy (OA), which measures the proportion of correctly classified pixels; Intersection over Union (IoU), assessing the overlap between predicted and ground truth areas; the F1 score, balancing precision and recall. The Kappa coefficient evaluates overall agreement, accounting for chance agreement. Additionally, the Separated Kappa (SeK) coefficient was utilized to address the class-imbalance issue, offering a more stringent assessment than the traditional Kappa coefficient[5].

*3.1.3 Benchmark methods:*
A range of deep learning-based methods from recent literature were selected for comparative evaluation with our proposed model. All methods underwent identical training protocols, ensuring an equitable and objective comparison.

**PCC**: the traditional method comparing classification results of two temporal images to detect semantic changes.

**HRSCD** [1]: The HRSCD dataset employs four SCD strategies: 1) PCC, using a UNet-like encoder-decoder FCN for change detection; 2) Directed classification, treating SCD as multi-temporal classification with bitemporal images; 3) Separated BCD and classification, combining binary change maps with classification maps for SCD; 4) Integrated BCD and classification, fusing bitemporal features from segmentation and BCD encoders to enhance SCD accuracy.

**ChangeMask** [3]: A deep multi-task encoder-transformer-decoder architecture designed for efficient and robust semantic change detection.

**SSESN** [6]: SSESN enhances semantic change detection by aggregating spatial and semantic information and employing a change-aware module for refined feature analysis and region categorization.

**Bi-SRNet** [7]: Bi-SRNet integrates semantic temporal features with reasoning blocks, enhancing segmentation of semantic categories and change detection accuracy.

### 3.2 Analysis of semantic change detection results
*3.2.1 Benchmark methods*
The qualitative results of different SCD methods on the Hi-UCD dataset is shown in **Fig. 2**. The PCC Deeplabv3+ method tends to over-segment the changed areas, as evidenced by the spread of green pixels beyond the ground truth. HRSCD_str4 shows a more conservative change detection, yet with some missed detections. SSESN offers a tighter alignment with the ground truth but still omits some minor changes. ChangeMask provides a closer approximation to the ground truth, with fewer omissions and misclassifications. MapChange demonstrates a superior delineation of changed areas, with a discernible reduction in false positives and negatives, closely matching the ground truth and displaying enhanced detection of complex change patterns.

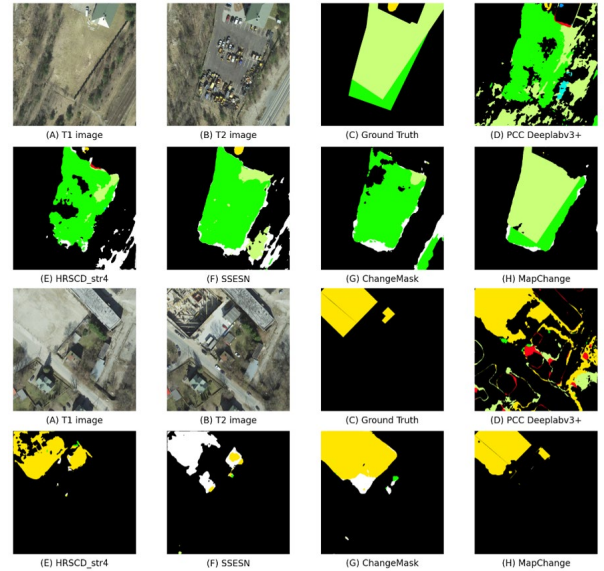

**Fig. 2.** Visualization results on the Hi-UCD dataset.

**Table I** and **Table II** Shows the quantitative assessment on the Hi-UCD and HRSCD datasets. Traditional PCC methods, utilizing various decoders like DeepLabv3+ and PSPNet, generally show lower SeK scores, indicating weakness of PCC-based methods. On the HRSCD dataset, PCC methods yield negative Kappa values, suggesting serious misclassification relative to the ground truth.

**Table I.** Quantitative results for the Hi-UCD dataset.

| Model | Decoder | Kappa(%) | Sek(%) | IoU(%) | F1(%) | OA(%) |
|---|---|---|---|---|---|---|
| PCC | Deeplabv3+ | 17.83 | 8.99 | 31.52 | 47.93 | 79.30 |
| PCC | U-Net++ | 24.16 | 12.81 | 36.53 | 53.52 | 82.95 |
| PCC | U-Net | 23.12 | 12.13 | 35.49 | 52.39 | 82.76 |
| PCC | FPN | 19.90 | 10.13 | 32.45 | 49.00 | 79.87 |
| PCC | PSPNet | 9.36 | 4.34 | 23.06 | 37.48 | 68.94 |
| hrscd_str1[1] | U-Net | 17.47 | 9.45 | 38.52 | 55.61 | 84.66 |
| hrscd_str2[1] | U-Net | -0.76 | -0.31 | 9.48 | 17.31 | 19.99 |
| hrscd_str3[1] | U-Net | 21.51 | 12.75 | 47.65 | 64.55 | 90.34 |
| hrscd_str4[1] | U-Net | 26.44 | 16.57 | 53.28 | 69.52 | 91.95 |
| SSESN[6] | SSFA+CA[6] | 16.04 | 10.02 | 52.95 | 69.23 | 90.62 |
| Bi-SRNet[7] | SR[7] | 14.76 | 9.00 | 50.51 | 67.12 | 90.80 |
| ChangeMask[3] | U-Net | 33.39 | 22.46 | 60.35 | 75.27 | 92.63 |
| MapChange | U-Net | **46.15** | **32.10** | **63.71** | **77.83** | **94.27** |

Advanced methods like ChangeMask and Bi-SRNet show improved metrics, with ChangeMask demonstrating respectable performance, particularly in IoU and F1 scores on the Hi-UCD dataset. However, MapChange consistently leads, suggesting that its approach to integrating temporal and semantic information effectively addresses SCD's inherent challenges. The trend across datasets indicates that methods leveraging deep feature integration and sophisticated architectural designs, such as MapChange, can better manage the complexity of high-resolution SCD tasks. These findings highlight a shift towards more integrated and contextually aware SCD models to improve detection accuracy and reliability.

**Table II.** Quantitative results for the HRSCD dataset.

| Model | Decoder | Kappa(%) | Sek(%) | IoU(%) | F1(%) | OA(%) |
|---|---|---|---|---|---|---|
| PCC | Deeplabv3+ | -5.43 | -2.86 | 35.90 | 52.84 | 79.34 |
| PCC | U-Net++ | -2.92 | -1.54 | 35.90 | 52.83 | 80.09 |
| PCC | U-Net | -6.42 | -3.34 | 34.68 | 51.50 | 79.02 |
| PCC | FPN | -2.64 | -1.39 | 36.02 | 52.96 | 79.30 |
| PCC | PSPNet | -6.92 | -3.57 | 33.61 | 50.32 | 78.02 |
| hrscd_str1[1] | U-Net | -5.94 | -3.03 | 32.83 | 49.44 | 79.28 |
| hrscd_str2[1] | U-Net | 9.38 | 4.20 | 19.70 | 32.92 | 13.74 |
| hrscd_str3[1] | U-Net | 1.40 | 0.82 | 46.47 | 63.45 | 80.59 |
| hrscd_str4[1] | U-Net | -0.57 | -0.32 | 44.49 | 61.58 | 83.19 |
| SSESN[6] | SSFA+CA[6] | -2.36 | -1.25 | 36.70 | 53.70 | 81.79 |
| Bi-SRNet[7] | SR[7] | 6.39 | 3.88 | 50.15 | 66.80 | 86.12 |
| ChangeMask[3] | U-Net | 2.93 | 1.63 | 41.22 | 58.38 | 85.32 |
| MapChange | U-Net | **16.30** | **10.50** | **56.05** | **71.83** | **89.07** |

*3.3.2 Comparison between modal fusion operations*

Experimental results for modal fusion operations in the MapChange framework are presented in Table II, contrasting the effectiveness of various fusion strategies between change and map features. While the Temporal-Symmetric Transformer (TST) was originally designed for symmetrical image comparison, the adaptation of TST-Add for fusion with map features yields suboptimal results. In contrast, simpler fusion operations such as concatenation (Cat) and addition (Add) demonstrate superior performance. Concatenation achieves the highest metrics. These findings indicate that straightforward operations like concatenation and addition are more effective for integrating change and map features within the MapChange framework.

**Table III.** Comparison between different modal fusion operation for the Hi-UCD dataset.

| Method | Kappa(%) | Sek(%) | IoU(%) | F1(%) | OA(%) |
|---|---|---|---|---|---|
| TST-Add | 35.85 | 21.14 | 47.16 | 64.10 | 93.67 |
| Add | 46.08 | 32.01 | 63.56 | 77.72 | 94.20 |
| Cat | 46.15 | 32.10 | 63.71 | 77.83 | 94.27 |

## 4. CONCLUSION

This paper addressed existing gaps in semantic change detection, notably the limitations of image pair comparison methods that struggle with temporal variance and minor change detection. Our contribution, the MapChange framework, introduce temporal invariant historical map data to provide semantic priors, enhancing change detection against these challenges. The innovative triplet network architecture within MapChange effectively processes multi-modal inputs, yielding a significant performance uplift. Experimental results on the Hi-UCD and HRSCD datasets demonstrate MapChange's superior performance.